# The Evolution of Football Betting: A Machine Learning Approach to Match Outcome Forecasting and Bookmaker Odds Estimation

Purnachandra Mandadapu
mpchandra39@gmail.com
Deloitte, Dallas, Texas, USA

**Abstract**

This paper explores the significant history of professional football and the betting industry, tracing its evolution from clandestine beginnings to a lucrative multi-million-pound enterprise. Initiated by the legalization of gambling in 1960 and complemented by advancements in football data gathering pioneered by Thorold Charles Reep, the symbiotic relationship between these sectors has propelled rapid growth and innovation. Over the past six decades, both industries have undergone radical transformations, with data collection methods evolving from rudimentary notetaking to sophisticated technologies such as high-definition cameras and Artificial Intelligence (AI)-driven analytics. Therefore, the primary aim of this study is to utilize Machine Learning (ML) algorithms to forecast premier league football match outcomes. By analyzing historical data and investigating the significance of various features, the study seeks to identify the most effective predictive models and discern key factors influencing match results. Additionally, the study aims to utilize these forecasting to inform the establishment of bookmaker odds, providing insights into the impact of different variables on match outcomes. By highlighting the potential for informed decision-making in sports forecasting and betting, this study opens up new avenues for research and practical applications in the domain of sports analytics.

## 1. Introduction

Since its inception in 19th-century England, professional football has shared an intricate relationship with the world of betting, initially existing in a legal gray area. What began as a mere enhancement to the viewing experience, injecting excitement and stakes into matches, soon morphed into a pivotal aspect of the sport's narrative—ushering some to glory while leading others to their downfall. The pivotal turning point arrived with the full legalization and standardization of gambling, courtesy of the British Act of Parliament known as "The Betting and Gaming Act" enacted on September 1st, 1960 [1]. This landmark legislation coincided with the burgeoning interest in football data gathering spearheaded by Thorold Charles Reep starting in 1951. The convergence of these events propelled both realms into widespread popularity and rapid advancement. Legalization provided the impetus to refine methods of data collection for football matches, crucial for supporting bookmakers in setting odds accurately. Over the subsequent six decades, both the betting and football data sectors underwent a remarkable metamorphosis. What began as humble betting establishments cautiously emerging from the shadows evolved into multi-million-pound conglomerates, boasting a global footprint and employing thousands worldwide. The era of gentlemen like Mr. Reep jotting down pass counts on scraps of paper transitioned into a lucrative industry utilized not only by betting entities but also by football clubs themselves. Today, the landscape of football data collection has undergone a seismic shift, transcending the rudimentary days of manual note-taking. The advent of cutting-edge technologies has birthed highly sophisticated tools for data aggregation, transforming every moment of a football match into a potential goldmine of insightful information. This includes an array of innovations such as multiple high-definition cameras meticulously tracking player movements, sensors embedded within players' boots capturing intricate details of their performance, and even microchips embedded within the ball itself to measure its trajectory and velocity. These advancements, coupled with myriad other data collection methodologies, offer a remarkably granular depiction of every match scenario.

Furthermore, the integration of Artificial Intelligence (AI) techniques, particularly Machine Learning (ML) [2]–[6], has expanded the horizons and capabilities of football data analysis. Extensive research has showcased the efficacy of ML in evaluating various facets of a football match. These insights serve as invaluable tools for fine-tuning player positioning, optimizing team strategies, refining training regimens, and, notably, predicting match outcomes—an area of keen interest driving the focus of this study. Within the ever-evolving realm of sports analytics, the intricate tapestry of football data stands as a reservoir ripe with untapped potential for making informed predictions. Therefore, this study endeavors not only to unlock this potential but also to construct a model adept at accurately prognosticating premier league football match outcomes. At the heart of this research lies a methodological approach that hinges upon the utilization of historical football data to fuel a spectrum of ML models. The aim is twofold: to scrutinize the relevance of various features within this data and to discern the optimal model for predicting match outcomes with precision. Furthermore, this study aims to utilize the forecasts generated by these models to recreate the process of establishing bookmaker odds, thus presenting a fresh perspective on calculating 1x2 bet odds. These bookmaker odds serve a dual function within this investigation. Firstly, they serve as a benchmark against which the predictions of ML models are evaluated. Secondly, they offer invaluable insights into the multifaceted influence of different factors on the outcome of a match.

This paper is as follows in the next section we will see the related works. In Section 3, the materials and methods are presented including data analysis. In Section 4, the experimental analysis is conducted. In Section 5, the results of the analysis are presented. In Section 6, the detailed discussion section is presented and we conclude the paper with some conclusion and future works in Section 7.

## 2. Related Works

This section discusses prior research related to forecasting football league match outcomes based on player characteristics and skills. Several studies have examined various aspects of this topic, providing valuable insights into predictive modeling techniques and their applications in sports betting such as [7] have noted limited support for the hypothesis of efficient betting markets, while [8] have shown rapid and complete adjustment of prices, especially after significant events like goals. Moreover, [9] has revealed that the number of fans supporting each club influences betting odds, favoring more popular teams. [10] has highlighted the potential for guaranteed positive returns by combining data sources. Additionally, [11], [12] have assessed the prediction accuracy of experts and betting exchange markets, offering insights into market behavior. Notably, [13] investigations into the behavior of publicly traded sports teams' stocks following match results have shown that abnormal returns for winning teams may be driven by investor sentiment rather than rational expectations. Furthermore, [14] analyses of the differences between sports betting and financial markets have indicated that bookmakers can outperform traditional brokers due to their ability to adjust prices dynamically.

Several research studies have [15]–[21] focused on predicting the outcomes of major sporting events, particularly the FIFA World Cup and the UEFA Euro Cup. A notable contribution was made by [22], who introduced a least squares betting approach in 1980 and applied it to data from the FIFA World Cup 1976. Additionally, [23] described a betting strategy based on the Fibonacci sequence, demonstrating its potential economic profitability during simulations of FIFA World Cup finals. [24] conducted an empirical study comparing prediction accuracy for the FIFA World Cup 2006 with predictions derived from the FIFA world ranking, highlighting the superiority of prediction markets. Moreover, [25] proposed a probabilistic prediction method for the 2018 FIFA World Cup based on the bookmaker consensus model, which was also employed for predicting the winner of the UEFA Euro Cup two years prior. Apart from the FIFA World Cup, [22] extended their forecasting approach to other sporting events like American football and basketball. Furthermore, [26] utilized a Bayesian dynamic generalized linear model to analyze team skills in the English premier league and the Spanish primera division, achieving impressive returns. However, existing literature lacks research on incorporating player attributes and skills into match outcome predictions for major football leagues and their second-tier counterparts. This study aims to address this gap by conducting a comprehensive analysis in this area.

## 3. Materials and Methods

In the realm of sports betting, the practice of bookmaking revolves around establishing odds for various outcomes of a sporting event. Bookmakers, also known as "bookies", undertake this task with the aim of attracting bets on all potential outcomes, thereby ensuring a profit regardless of the event's result. Odds are determined based on both statistical analysis involving complex algorithms and the subjective assessments of experts in the field. Lower odds indicate a higher probability of a specific outcome. For instance, if Team $A$ is perceived as stronger than Team $B$, the odds for Team $A$'s victory will be lower. A fundamental type of bet, known as "1x2" or "win-draw-win", encompasses three potential outcomes:

- "1" signifies a victory for the Home Team.
- "X" denotes a draw, and;
- "2" represents a triumph for the Away Team.

It is noteworthy that in most football leagues, each team competes against all others twice—once at home and once away. The venue significantly influences predictions, as it is well-documented in [27] that teams typically perform better in front of their home crowd. Odds, expressed as decimal numbers greater than 1, are calculated based on the perceived likelihood of an outcome occurring, although the sum of all probabilities may exceed 100%. This excess, termed the "overround" or "vigorish" ("vig"), guarantees profit for the bookmaker. The odds formula is represented as in Equation (1).

$$Odds = \frac{1}{P} \tag{1}$$

where "*Odds*" denotes the odds for the event, and "*P*" represents the probability of the event in decimal form. For instance, if the probability of Team *A* winning is 0.5 (50%), the odds would be computed as 1/0.5 = 2.00. This indicates that for every unit of currency wagered, the payout would be double if Team *A* emerges victorious. Notably, odds can vary between different bookmakers, influenced by algorithms and subjective expert opinions. This discrepancy becomes particularly significant in matches between evenly matched teams, where different betting companies may favor different outcomes. Furthermore, odds fluctuate dynamically leading up to an event due to various factors such as player injuries, alterations in team line-ups, and even punter behavior. However, once a bet is placed, the odds are fixed, safeguarding both the bettor and the bookmaker from subsequent changes. This presents an opportunity for strategizing, aiming to predict when odds for a desirable outcome will be highest and placing bets accordingly. However, akin to the adage "the house always wins", it is evident that bookmaking companies structure their operations to ensure profitability. Strategies such as favoring odds in their favor, imposing limits on maximum bet sizes, adjusting odds based on betting patterns, and at times, declining bets from successful bettors, have garnered controversy and criticism. While this study primarily concentrates on the 1x2 bet, it is essential to acknowledge that this is merely one among numerous bets offered by betting companies. Other examples include over/under bets, correct score bets, first goal scorer bets, yellow card bets, and a myriad of others.

**3.1 Data Analysis**

In this section, we'll delve into the dataset that's central to our research. This dataset includes detailed stats from the English premier league seasons of 2021-2022 and 2022-2023 [fbref.com]. It's packed with lots of different stats about matches, giving us a strong base for exploring which match features can help us predict outcomes. To get the data, we used a technique called web scraping, which helps collect lots of information from websites that don't offer easy ways to download their data. For our study, we used web scraping to gather match data for all the teams that played in the 2021-22 and 2022-2023 seasons. Our program was set up to visit the pages for each of the 20 teams, find the stats we were interested in, and organize them neatly into a database. This database became the foundation for our research. This study centers on extracting actionable data from the premier league table, encompassing statistics for each team across a 380-match season. Data is sourced from 9 sections including scores, shooting, goalkeeping, passing, and others. Each match's data for both teams is amalgamated into a dataset using shared columns. From each section, 34 pertinent statistics, alongside descriptive columns, are selected, resulting in a comprehensive table with 1520 rows and 52 columns. This dataset is primed for ML analysis, aiming to discern patterns and make predictions regarding match outcomes.

**3.1.1 Data Preprocessing**

Before starting with ML [28]–[33], it's crucial to prep the data. This involves cleaning up the raw data, making it ready for analysis. Missing data can be filled using methods like default filling, mean, median, or mode filling, or through prediction using algorithms like K Nearest Neighbour (KNN) or regression. Noisy data, which includes random variance or errors, can be addressed through techniques like binning, regression, or clustering. However, bringing together data from different sources to give users a unified view. But, this may lead to data redundancy, which needs careful handling. It requires to adjust the data to a suitable format for mining. This may involve normalization, aggregation, or generalization. Large datasets can be trimmed down while maintaining analytical results. Techniques for data reduction include reducing dimensionality, numerosity, or compression. Similarly, converting continuous attributes into categorical ones can speed up the mining process, although some information might be lost in the process. ML algorithms require numeric values, so encoding was necessary:

- The "venue" column was encoded with numbers: Home as 1 and Away as 0.
- The "opponent" column contained the names of away/opposition teams, each assigned an integer.
- Similarly, the "team" column had the names of home teams, each also assigned an integer.
- The "result" column, crucial for match outcomes, was mapped: Wins as 1 (W), Draws as 0 (D), and Losses as 2 (L), reflecting the 1x2 bet.

Additionally, irrelevant columns like "match report", "notes", "referee", "captain", and "formation" were removed. Specific preparations were made for ML algorithms. Initially, a normalized copy of the dataset was created for algorithms to function correctly. Finally, the last matchweek of the 2022-2023 season data was replaced with season-long averages for each team.

## 4. Experimental Analysis

This section delves into the technical aspects and methodology used in this study. The experimental evaluation aims to address the research questions—Different machine learning models offer distinct approaches to predicting outcomes, each with unique strengths and weaknesses. Therefore, proper feature and hyperparameter selection are crucial given the complexity of the data. This study seeks to compare the performance of various ML models, such as random forest and KNN, based on multiple metrics. Features in ML models represent measurable properties or characteristics of the analyzed observations. Understanding the importance of these features and how each model evaluates them is essential. Feature importance will be determined based on scores provided by the ML model's built-in metrics. The impact of features directly related to outcomes, such as goals scored or conceded in football, will be examined. Proper selection of training data is critical, particularly with time-related data such as football match histories. The dataset consists of matches from two premier league seasons, and various splitting methods will be applied to ensure accurate predictions. Model predictions will be translated into betting outcomes to assess their effectiveness. The accuracy of predictions will be evaluated based on feature and hyperparameter selection, as well as resistance to biases and competency in handling outlier outcomes. The selection of Python programming language for this research stemmed from its user-friendly structure and easy-to-understand syntax, facilitating the creation of a replicable and modifiable scheme. Jupyter Notebook served as the chosen development environment due to its interactive capabilities, allowing seamless integration of code, visualizations, and text for comprehensive data exploration and analysis.

Initially, the dataset consisted of numerous columns detailing each match's attributes. To ensure efficient processing by the ML algorithms employed in this study, it was necessary to reduce the number of parameters provided to these algorithms [34]–[37]. This task was accomplished using feature selectors. It's noteworthy that various selectors can be utilized for this purpose, and some are specific to particular algorithms. For this study, it was decided to employ a common selector, Recursive Feature Elimination (RFE), for all algorithms, supplemented by algorithm-specific selectors for comparison purposes. RFE is a method used to select features across iterations, starting with the entire set and recursively eliminating less important features until the desired number of features is reached. Another important aspect is the correlation matrix, a tabular representation displaying correlation coefficients between variables. This matrix aids in summarizing data and identifying relationships between variables. In the realm of ML, hyperparameters play a critical role in controlling the learning process. Unlike model parameters learned during training, hyperparameters are set prior to the learning process. Techniques such as grid search or randomized search were employed to systematically construct and assess models for various combinations of algorithm parameters specified within a defined grid or range. This systematic approach facilitates fine-tuning of ML models to enhance their performance. Evaluation metrics are indispensable for analyzing the accuracy of model predictions. The methods utilized in this study are derived from the sklearn library. The evaluation metrics include:

- **Accuracy Score**: A simple metric indicating the ratio of correct predictions to the total number of predictions.
- **Confusion Matrix:** A matrix depicting the classification performance, particularly useful in multiclass scenarios.
- **Classification Report:** A detailed breakdown of precision, recall, and F1-score for each class in multiclass classification tasks, along with overall metrics such as support, micro-average, macro-average, and weighted-average.

## 5. Result Analysis

In this study, the outcomes of the investigation aiming to explore and assess the effectiveness of various ML models will be delineated. The inquiry into this matter was systematically structured into a three-phase procedure. Initially, models were constructed for data partitions as delineated in Section 4. This methodology facilitates a deeper comprehension of each model's performance across different temporal frames and varying data compositions. Subsequently, the examination entailed testing the models utilizing diverse sets of features. This investigation was pivotal in comprehending the influence of specific variables on the predictive capabilities of the model. Finally, the trained models were leveraged to predict forthcoming match results. In the subsequent sections, intricate details of each phase will be expounded upon, furnishing a comprehensive overview of the obtained results and comparisons between the employed ML models.

### 5.1 Random Forest Results

In the evaluation of our model's performance across different splits of the dataset as shown in Table I, we employed various metrics to assess its accuracy and reliability. For the two-season split, the random forest classifier demonstrated an accuracy of 64.95%, with notable precision and recall rates for different classes. However, there was a discernible bias towards certain classes, particularly evident in misclassification patterns. Transitioning to a one-season split yielded a slight improvement in overall accuracy to 67.33%, albeit with persistent bias towards specific classes. Conversely, for the recent matches split, accuracy notably decreased to 47.73%, mainly due to challenges in classifying one particular class, indicating limitations in prediction based solely on recent data.

TABLE I
RANDOM FOREST CLASSIFIER RESULTS FOR PREDICTIONS IN MULTIPLE TIME SPLITS

| Random Forest Classifier | Accuracy | Class | Precision | Recall | F-1 Score |
|---|---|---|---|---|---|
| 2 Seasons of Data | 0.64 | 0 | 0.34 | 0.17 | 0.23 |
| | | 1 | 0.72 | 0.75 | 0.74 |
| | | 2 | 0.66 | 0.83 | 0.74 |
| 1 Season of Data | 0.67 | 0 | 0.29 | 0.06 | 0.10 |
| | | 1 | 0.66 | 0.81 | 0.73 |
| | | 2 | 0.73 | 0.86 | 0.79 |
| 10 Match Weeks of Data | 0.47 | 0 | 0 | 0 | 0 |
| | | 1 | 0.56 | 0.71 | 0.63 |
| | | 2 | 0.42 | 0.79 | 0.55 |

In exploring subsets of features, our analysis uncovered intriguing insights (refer to Table II). The first subset encompassing all available features exhibited a moderate accuracy of 68%, showcasing a balanced performance across classes. Interestingly, employing feature selection techniques like RFE led to a marginal enhancement in accuracy to 69%, underscoring the significance of relevant feature inclusion. However, selecting features based solely on their correlation with the target variable resulted in decreased accuracy (62%), highlighting the limitations of correlation as a sole criterion for feature selection.

TABLE II
RFE RESULTS FOR MULTIPLE SETS OF FEATURES

| Random Forest Classifier | Accuracy | Class | Precision | Recall | F-1 Score |
|---|---|---|---|---|---|
| All Feature Subset | 0.68 | 0 | 0.43 | 0.28 | 0.34 |
| | | 1 | 0.68 | 0.78 | 0.72 |
| | | 2 | 0.77 | 0.80 | 0.78 |
| RFE Feature Subset | 0.68 | 0 | 0.42 | 0.41 | 0.41 |
| | | 1 | 0.75 | 0.76 | 0.76 |
| | | 2 | 0.76 | 0.76 | 0.76 |
| Co-relation Feature Subset | 0.62 | 0 | 0.30 | 0.31 | 0.31 |
| | | 1 | 0.67 | 0.68 | 0.64 |
| | | 2 | 0.75 | 0.73 | 0.74 |

Additionally, our model's predictive capabilities regarding football match outcomes were scrutinized. Notably, the model displayed strong inclinations towards certain outcomes, reflecting inherent uncertainties in football matches (refer to Table III). For instance, matches involving teams like Leeds United and Tottenham exhibited pronounced biases, indicating the model's confidence in predicting certain outcomes. Conversely, matches such as Crystal Palace vs Nottingham Forest showcased a strong inclination towards a draw, indicative of the model's ability to capture uncertainty and variability in football predictions. This nuanced approach not only enhances prediction accuracy but also provides valuable insights into the inherent unpredictability of football matches.

TABLE III
MATCHWEEK 38 GAMES FORECASTED RESULTS FOR RANDOM FOREST CLASSIFIER

| Home Team | AwayTeam | Real result | Predicted 0 | Predicted 1 | Predicted 2 |
|---|---|---|---|---|---|
| Crystal Palace | Nottingham sorest | 0 | 3833 | 1096 | 66 |
| Leeds United | Tottenham | 2 | 1016 | 0 | 3934 |
| Arsenal | 'Outlives | 1 | 568 | 3997 | 435 |
| Leicester City | West Ham | 1 | 3440 | 3 | 1557 |
| Manchester United | Fulham | 1 | 1345 | 3655 | 0 |
| Everton | Bournemouth | 1 | 3336 | 4.57 | 1207 |
| Aston Villa | Brighton | 1 | 390 | 1911 | 2199 |
| Chelsea | Newcastle United | 0 | 2461 | 37 | 2502 |

**5.2 Support Vector Machine Results**

In this section, we evaluate the performance of a Support Vector Machine (SVM) model across different datasets (refer to Table IV). For the 2-season dataset, the SVM model achieves an accuracy of 67%, showing strengths in predicting "away win" outcomes but displaying limitations in correctly identifying draws, with only 30 out of 92 instances accurately predicted. Similar trends are observed in the 1-season dataset, where the accuracy improves to 72.67%, yet the model still exhibits challenges in predicting draw outcomes. Notably, when applied to recent matches, the SVM model's performance notably declines to 45%, indicating difficulties in predicting match results, possibly due to the smaller dataset size and increased volatility. When utilizing all features, the SVM model attains an accuracy of 72%, performing relatively well in predicting "home" and "away" wins but struggling with draw predictions. RFE slightly reduces the accuracy to 70%, indicating that while effective in feature selection, it doesn't significantly enhance draw prediction.

Further, using only the best-correlated features results in a decreased accuracy of 66.67%, emphasizing that while correlation aids in feature selection, it may overlook crucial relationships for draw prediction.

TABLE IV
SUPPORT VECTOR MACHINE RESULTS FOR PREDICTIONS IN MULTIPLE TIME SPLITS

| Support Vector Machine | Accuracy | Class | Precision | Recall | F-1 Score |
|---|---|---|---|---|---|
| 2 Seasons of Data | 0.66 | 0 | 0.37 | 0.33 | 0.34 |
|  |  | 1 | 0.79 | 0.74 | 0.76 |
|  |  | 2 | 0.71 | 0.80 | 0.75 |
| 1 Season of Data | 0.72 | 0 | 0.50 | 0.34 | 0.41 |
|  |  | 1 | 0.76 | 0.81 | 0.79 |
|  |  | 2 | 0.77 | 0.85 | 0.81 |
| 10 Match Weeks of Data | 0.45 | 0 | 0.60 | 0.19 | 0.29 |
|  |  | 1 | 0.41 | 0.64 | 0.50 |
|  |  | 2 | 0.47 | 0.57 | 0.52 |

Despite generally decent results, the SVM model consistently faces challenges in predicting draws, suggesting potential complexities or overlooked features specific to draw outcomes (refer to Table V). To address this, a refined feature engineering and selection process could be beneficial, focusing on features pertinent to draw prediction. Notably, the model displays high confidence in certain predictions, yet errors in predicted outcomes, particularly for matches involving Leeds, Tottenham, Arsenal, Wolves, Chelsea, Newcastle, Southampton, and Liverpool, underscore the limitations of purely statistical analysis in match prediction as shown in Table VI. Additionally, the model's reluctance to settle on a third outcome for some matches suggests nuances in match dynamics or team strengths that warrant further investigation.

TABLE V
RFE RESULTS FOR MULTIPLE SETS OF FEATURES

| Support Vector Machine | Accuracy | Class | Precision | Recall | F-1 Score |
|---|---|---|---|---|---|
| All Feature Subset | 0.72 | 0 | 0.50 | 0.41 | 0.45 |
|  |  | 1 | 0.77 | 0.80 | 0.78 |
|  |  | 2 | 0.76 | 0.81 | 0.79 |
| RFE Feature Subset | 0.70 | 0 | 0.53 | 0.31 | 0.39 |
|  |  | 1 | 0.77 | 0.81 | 0.79 |
|  |  | 2 | 0.68 | 0.90 | 0.73 |
| Co-relation Feature Subset | 0.66 | 0 | 0.36 | 0.16 | 0.22 |
|  |  | 1 | 0.73 | 0.90 | 0.76 |
|  |  | 2 | 0.67 | 0.81 | 0.73 |

TABLE VI
MATCHWEEK 38 GAMES FORECASTED RESULTS FOR SUPPORT VECTOR MACHINE

| Home Team | AwayTeam | Real result | Predicted 0 | Predicted 1 | Predicted 2 |
|---|---|---|---|---|---|
| Crystal Palace | Nottingham Forest | 0 | 150 | 150 | 0 |
| Leeds United | Tottenham | 2 | 0 | 0 | 300 |
| Arsenal | Wolves | 1 | 0 | 300 | 0 |
| Leicester City | West Ham | 1 | 150 | 0 | 150 |
| Manchester United | 'Fulham | 1 | 150 | 150 | 0 |
| Everton | Bournemouth | 1 | 0 | 150 | 150 |
| Aston Villa | Brighton | 1 | 0 | 150 | 150 |
| Chelsea | Newcastle United | 0 | 0 | 0 | 300 |

**5.3 K-Nearest Neighbor Results**

In examining the performance of various ML models, particularly the KNN and SVM models, several key metrics were considered. Notably, the accuracy of the KNN model on 2-season data is observed to be 61.52%, slightly lower than that of the SVM model as shown in Table VII. Delving deeper, it's evident that the KNN model excels in predicting away wins, correctly identifying 125 out of 158 instances, whereas its performance is notably poorer in predicting draws, with only 9 correct predictions out of 92 instances. This trend persists even when focusing on 1-season data, although a slight improvement in accuracy to 62.67% is noted for the KNN model.

TABLE VII
KNN RESULTS FOR PREDICTIONS IN MULTIPLE TIME SPLITS

| KNN | Accuracy | Class | Precision | Recall | F-1 Score |
|---|---|---|---|---|---|
| 2 Seasons of Data | 0.62 | 0 | 0.25 | 0.03 | 0.06 |
|  |  | 1 | 0.60 | 0.83 | 0.70 |
|  |  | 2 | 0.58 | 0.75 | 0.71 |
| 1 Season of Data | | 0 | 0.30 | 0.19 | 0.23 |
| | 0.65 | 1 | 0.70 | 0.81 | 0.75 |
| | | 2 | 0.72 | 0.75 | 0.73 |
| 10 Match Weeks of Data | | 0 | 0.50 | 0.19 | 0.27 |
| | 0.68 | 1 | 0.73 | 0.81 | 0.77 |
| | | 2 | 0.58 | 0.33 | 0.75 |

However, a significant decline in accuracy to 38.64% is observed for recent matches when utilizing the KNN model (refer to Table VIII). This decline is mirrored in the precision, recall, and F1-score metrics across all classes, indicating suboptimal predictive performance. Similar challenges are encountered with the SVM model, which may be attributed to the reduced dataset size and heightened volatility of recent matches. Despite efforts to enhance performance through feature selection techniques like RFE, the KNN model's accuracy remains lower compared to SVM and random forest classifier, standing at 65.33%. Interestingly, the KNN model exhibits improved performance when trained on features selected via correlation matrix analysis.

TABLE VIII
RFE RESULTS FOR MULTIPLE SETS OF FEATURES

| KNN | Accuracy | Class | Precision | Recall | F-1 Score |
|---|---|---|---|---|---|
| All Feature Subset | | 0 | 0.25 | 0.03 | 0.06 |
| | 0.62 | 1 | 0.60 | 0.83 | 0.70 |
| | | 2 | 0.58 | 0.75 | 0.71 |
| RFE Feature Subset | | 0 | 0.30 | 0.19 | 0.23 |
| | 0.65 | 1 | 0.70 | 0.81 | 0.75 |
| | | 2 | 0.72 | 0.75 | 0.73 |
| Co-relation Feature Subset | | 0 | 0.50 | 0.19 | 0.27 |
| | 0.68 | 1 | 0.73 | 0.81 | 0.77 |
| | | 2 | 0.58 | 0.33 | 0.75 |

## 5.4 Extreme Gradient Boosting Results

The section presents findings regarding the performance of the Extreme Gradient Boost (XGB) model in comparison to other models. Across two seasons of data and one season of data, the accuracies remained consistently between 65% to 70% (refer to Table IX). Notably, similar to the KNN model, XGB demonstrated notable enhancement in accuracy, particularly when specific hyperparameters were selected.

TABLE IX
XGB RESULTS FOR PREDICTIONS IN MULTIPLE TIME SPLITS

| XGB | Accuracy | Class | Precision | Recall | F-1 Score |
|---|---|---|---|---|---|
| 2 Seasons of Data | | 0 | 0.43 | 0.25 | 0.32 |
| | 0.63 | 1 | 0.78 | 0.77 | 0.77 |
| | | 2 | 0.69 | 0.85 | 0.76 |
| 1 Season of Data | | 0 | 0.33 | 0.06 | 0.11 |
| | 0.68 | 1 | 0.74 | 0.81 | 0.77 |
| | | 2 | 0.67 | 0.90 | 0.77 |
| 10 Match Weeks of Data | | 0 | 0.44 | 0.25 | 0.32 |
| | 0.47 | 1 | 0.67 | 0.43 | 0.52 |
| | | 2 | 0.42 | 0.79 | 0.55 |

A noteworthy boost of approximately 6% in accuracy highlights the significance of meticulous feature and hyperparameter selection for this model (refer to Table X). Moreover, XGB emerged as one of the models capable of introducing essential variability to its outcomes. However, it exhibited a tendency towards polarization, particularly in comparison to the random forest classifier, where instances of the third class often exhibited significantly lower records than the other two classes. Additionally, intriguingly, instances were observed where classes were evenly distributed among matches, indicating a unique pattern within the data.

TABLE X
RFE RESULTS FOR MULTIPLE SETS OF FEATURES

| XGB | Accuracy | Class | Precision | Recall | F-1 Score |
|---|---|---|---|---|---|
| All Feature Subset | | 0 | 0.33 | 0.05 | 0.11 |
| | 0.63 | 1 | 0.74 | 0.81 | 0.77 |
| | | 2 | 0.67 | 0.90 | 0.77 |
| RFE Feature Subset | | 0 | 0.38 | 0.09 | 0.15 |
| | 0.66 | 1 | 0.75 | 0.81 | 0.78 |
| | | 2 | 0.63 | 0.33 | 0.72 |
| Co-relation Feature Subset | 0.66 | 0 | 0.20 | 0.16 | 0.23 |
| | | 1 | 0.71 | 0.75 | 0.73 |

| | | 2 | 0.66 | 0.35 | 0.74 |

## 6. Discussion

In this section, we shift our focus from objectively analyzing results to interpretatively exploring them. This provides an opportunity to reflect on the outcomes of our investigation, enabling the extraction of meaningful insights and formulation of valuable conclusions. Comparison of various ML models on a consistent dataset has unveiled distinct strengths and weaknesses inherent in each approach. The examination of four models under scrutiny has showcased diverse performance outcomes, underscoring the algorithmic diversity and their tailored approaches to addressing different classification tasks. Notably, SVM has emerged as the frontrunner, not only in terms of accuracy but also across other performance metrics. Random forest classifier and KNN have exhibited comparable levels of average precision and F1-score, with random forest classifier displaying a slight edge in recall. This observation hints at potential challenges faced by KNN, particularly in mitigating false negatives, potentially influenced by its sensitivity to neighbor selection and data dimensionality, while highlighting random forest classifier's enhanced generalization owing to its ensemble nature. Interestingly, despite similar accuracy to random forest classifier and KNN, the XGBoost model lags in other performance metrics. XGBoost's robustness is typically acknowledged; however, its performance might be subject to variations based on the choice of loss function and regularization parameters.

In summary, our findings underscore the multidimensional nature of model evaluation in ML. Each model offers distinctive capabilities, and the selection depends on the specific requirements of the prediction task. Feature selection emerges as a pivotal step in ML, directly influencing model performance. Our experiment compared model accuracy using various feature selection techniques, revealing nuances in performance across different models. While random forest demonstrated superior performance with features selected by RFE, SVM and XGBoost excelled with all features. Additionally, KNN exhibited superior prediction capabilities with features selected based on correlation. On average, models marginally outperformed subsets of features when employing all 34 features instead of subsets of 10 features chosen by selectors. However, the discernible differences were negligible, suggesting that there's no universally superior method for feature selection. Thus, while feature selection plays a significant role, it's imperative to explore diverse approaches tailored to different models. Furthermore, the size of subsets warrants further investigation. Moreover, features such as expected goals (*xg*), expected goals against (*xga*), shot-creating actions (*sca*), and goal-creating actions (*gca*) consistently emerged as the most influential, aligning with the fundamental strategies of football. This underscores the impact of feature characteristics on predictive success, highlighting the importance of further analyzing match statistics. Aligned with our ML assessment, we observed that the choice of the time window significantly influences prediction accuracy. The optimal balance between historical data and recent performance metrics remains pivotal for achieving optimal predictive accuracy.

## 7. Conclusion and Future Works

This study aims to evaluate various ML models using a specific dataset to determine their effectiveness in predicting football match outcomes. Some models exhibited strong performance with specific features, showcasing the adaptability of ML algorithms in this domain. A noteworthy aspect of our investigation is the exploration of Explainable Artificial Intelligence (XAI). The interpretability of ML models is crucial, especially in complex fields like football match prediction. Further analysis delved into feature importance, time window significance, and the accuracy of "1x2" odds calculated through ML. The variability of predictions based on data splits suggests optimization opportunities for handling recent data effectively. Additionally, refining algorithms to address edge cases and extreme values in "1x2" odds calculation is imperative. However, recognizing the field's evolution, we propose several enhancements such as expansion of the dataset to include more descriptive features and entries. Improved data preprocessing by consolidating related match data. Extensive exploration of hyperparameters. Investigation of varying feature subset sizes selected by RFE and Correlation. Integration of advanced ML algorithms, potentially including neural networks. Increased simulation experiments to explore various scenarios comprehensively. Considering alternative model targets, such as goals scored and conceded, to enhance prediction accuracy. Exploring diverse betting strategies to complement predictive models. Lastly, while our study aids in enhancing betting strategies, it's crucial to understand that models alone cannot consistently outperform bookmakers. Success lies in balancing data analysis with domain expertise.

## References

[1] B. Harris, "Federal Interference with State and Tribal Sports Betting Regulations Will Not Work: Where the Sports Wagering Integrity Act of 2018 Went Wrong and How Federal Legislation Might Be Effective," *Journal of Legal Aspects of Sport*, vol. 30, no. 2, pp. 106–141, Aug. 2020, doi: 10.18060/24253.

[2] V. K. Kanaparthi, "Examining the Plausible Applications of Artificial Intelligence & Machine Learning in Accounts Payable

Improvement," *FinTech*, vol. 2, no. 3, pp. 461–474, Jul. 2023, doi: 10.3390/fintech2030026.

[3] V. Kanaparthi, "Examining Natural Language Processing Techniques in the Education and Healthcare Fields," *International Journal of Engineering and Advanced Technology*, vol. 12, no. 2, pp. 8–18, Dec. 2022, doi: 10.35940/ijeat.b3861.1212222.

[4] V. Kanaparthi, "Credit Risk Prediction using Ensemble Machine Learning Algorithms," in *6th International Conference on Inventive Computation Technologies, ICICT 2023 - Proceedings*, 2023, pp. 41–47. doi: 10.1109/ICICT57646.2023.10134486.

[5] V. Kanaparthi, "AI-based Personalization and Trust in Digital Finance," Jan. 2024, Accessed: Feb. 04, 2024. [Online]. Available: https://arxiv.org/abs/2401.15700v1

[6] V. Kanaparthi, "Exploring the Impact of Blockchain, AI, and ML on Financial Accounting Efficiency and Transformation," Jan. 2024, Accessed: Feb. 04, 2024. [Online]. Available: https://arxiv.org/abs/2401.15715v1

[7] R. G. Ricard Gil and S. D. Levitt, "TESTING THE EFFICIENCY OF MARKETS IN THE 2002 WORLD CUP," *The Journal of Prediction Markets*, vol. 1, no. 3, pp. 255–270, Dec. 2012, doi: 10.5750/jpm.v1i3.504.

[8] K. Croxson and J. J. Reade, "Information and efficiency: Goal arrival in soccer betting," *Economic Journal*, vol. 124, no. 575, pp. 62–91, Mar. 2014, doi: 10.1111/ecoj.12033.

[9] D. Forrest and R. Simmons, "Sentiment in the betting market on Spanish football," *Applied Economics*, vol. 40, no. 1, pp. 119–126, Jan. 2008, doi: 10.1080/00036840701522895.

[10] E. Franck, E. Verbeek, and S. Nüesch, "Prediction accuracy of different market structures - bookmakers versus a betting exchange," *International Journal of Forecasting*, vol. 26, no. 3, pp. 448–459, Jul. 2010, doi: 10.1016/j.ijforecast.2010.01.004.

[11] M. Spann and B. Skiera, "Sports forecasting: A comparison of the forecast accuracy of prediction markets, betting odds and tipsters," *Journal of Forecasting*, vol. 28, no. 1, pp. 55–72, Jan. 2009, doi: 10.1002/for.1091.

[12] H. O. Stekler, D. Sendor, and R. Verlander, "Issues in sports forecasting," *International Journal of Forecasting*, vol. 26, no. 3, pp. 606–621, Jul. 2010, doi: 10.1016/j.ijforecast.2010.01.003.

[13] F. Palomino, L. Renneboog, and C. Zhang, "Information salience, investor sentiment, and stock returns: The case of British soccer betting," *Journal of Corporate Finance*, vol. 15, no. 3, pp. 368–387, Jun. 2009, doi: 10.1016/j.jcorpfin.2008.12.001.

[14] S. D. Levitt, "Why are gambling markets organised so differently from financial markets?," *Economic Journal*, vol. 114, no. 495, pp. 223–246, Apr. 2004, doi: 10.1111/j.1468-0297.2004.00207.x.

[15] G. S. Kashyap, K. Malik, S. Wazir, and R. Khan, "Using Machine Learning to Quantify the Multimedia Risk Due to Fuzzing," *Multimedia Tools and Applications*, vol. 81, no. 25, pp. 36685–36698, Oct. 2022, doi: 10.1007/s11042-021-11558-9.

[16] G. S. Kashyap *et al.*, "Detection of a facemask in real-time using deep learning methods: Prevention of Covid 19," Jan. 2024, Accessed: Feb. 04, 2024. [Online]. Available: https://arxiv.org/abs/2401.15675v1

[17] G. S. Kashyap, A. E. I. Brownlee, O. C. Phukan, K. Malik, and S. Wazir, "Roulette-Wheel Selection-Based PSO Algorithm for Solving the Vehicle Routing Problem with Time Windows," Jun. 2023, Accessed: Jul. 04, 2023. [Online]. Available: https://arxiv.org/abs/2306.02308v1

[18] G. S. Kashyap, A. Siddiqui, R. Siddiqui, K. Malik, S. Wazir, and A. E. I. Brownlee, "Prediction of Suicidal Risk Using Machine Learning Models." Dec. 25, 2021. Accessed: Feb. 04, 2024. [Online]. Available: https://papers.ssrn.com/abstract=4709789

[19] G. S. Kashyap, D. Mahajan, O. C. Phukan, A. Kumar, A. E. I. Brownlee, and J. Gao, "From Simulations to Reality: Enhancing Multi-Robot Exploration for Urban Search and Rescue," Nov. 2023, Accessed: Dec. 03, 2023. [Online]. Available: https://arxiv.org/abs/2311.16958v1

[20] G. S. Kashyap *et al.*, "Revolutionizing Agriculture: A Comprehensive Review of Artificial Intelligence Techniques in Farming," Feb. 2024, doi: 10.21203/RS.3.RS-3984385/V1.

[21] P. Kaur, G. S. Kashyap, A. Kumar, M. T. Nafis, S. Kumar, and V. Shokeen, "From Text to Transformation: A Comprehensive Review of Large Language Models' Versatility," Feb. 2024, Accessed: Mar. 21, 2024. [Online]. Available: https://arxiv.org/abs/2402.16142v1

[22] R. Stefani, "Improved Least Squares Football, Basketball, and Soccer Predictions," *IEEE Transactions on Systems, Man, and Cybernetics*, vol. 10, no. 2, pp. 116–123, Jul. 2008, doi: 10.1109/tsmc.1980.4308442.

[23] F. Archontakis and E. Osborne, "Playing It Safe? A Fibonacci Strategy for Soccer Betting," *Journal of Sports Economics*, vol. 8, no. 3, pp. 295–308, Jun. 2007, doi: 10.1177/1527002506286775.

[24] S. Luckner, J. Schröder, and C. Slamka, "On the Forecast Accuracy of Sports Prediction Markets," in *Negotiation, Auctions, and Market Engineering*, Springer, Berlin, Heidelberg, 2008, pp. 227–234. doi: 10.1007/978-3-540-77554-6_17.

[25] A. Zeileis, C. Leitner, and K. Hornik, "Probabilistic forecasts for the eeee FIFA World Cup based on the bookmaker consensus model," *Working Papers in Economics and Statistics*, 2018, Accessed: Mar. 21, 2024. [Online]. Available: https://www.econstor.eu/handle/10419/184987

[26] H. Rue and Ø. Salvesen, "Prediction and retrospective analysis of soccer matches in a league," *Journal of the Royal Statistical Society Series D: The Statistician*, vol. 49, no. 3, pp. 399–418, Sep. 2000, doi: 10.1111/1467-9884.00243.

[27] "Premier League percentage of points (home and away)." https://www.soccerstats.com/table.asp?league=england&tid=1 (accessed Mar. 21, 2024).

[28] H. Habib, G. S. Kashyap, N. Tabassum, and T. Nafis, "Stock Price Prediction Using Artificial Intelligence Based on LSTM–Deep Learning Model," in *Artificial Intelligence & Blockchain in Cyber Physical Systems: Technologies & Applications*, CRC Press, 2023, pp. 93–99. doi: 10.1201/9781003190301-6.

[29] S. Wazir, G. S. Kashyap, and P. Saxena, "MLOps: A Review," Aug. 2023, Accessed: Sep. 16, 2023. [Online]. Available: https://arxiv.org/abs/2308.10908v1

[30] S. Naz and G. S. Kashyap, "Enhancing the predictive capability of a mathematical model for pseudomonas aeruginosa through artificial neural networks," *International Journal of Information Technology 2024*, pp. 1–10, Feb. 2024, doi: 10.1007/S41870-023-01721-W.

[31] N. Marwah, V. K. Singh, G. S. Kashyap, and S. Wazir, "An analysis of the robustness of UAV agriculture field coverage using multi-agent reinforcement learning," *International Journal of Information Technology (Singapore)*, vol. 15, no. 4, pp. 2317–2327, May 2023, doi: 10.1007/s41870-023-01264-0.

[32] S. Wazir, G. S. Kashyap, K. Malik, and A. E. I. Brownlee, "Predicting the Infection Level of COVID-19 Virus Using Normal Distribution-Based Approximation Model and PSO," Springer, Cham, 2023, pp. 75–91. doi: 10.1007/978-3-031-33183-1_5.

[33] M. Kanojia, P. Kamani, G. S. Kashyap, S. Naz, S. Wazir, and A. Chauhan, "Alternative Agriculture Land-Use Transformation Pathways by Partial-Equilibrium Agricultural Sector Model: A Mathematical Approach," Aug. 2023, Accessed: Sep. 16, 2023. [Online]. Available: https://arxiv.org/abs/2308.11632v1

[34] V. Kanaparthi, "Evaluating Financial Risk in the Transition from EONIA to ESTER: A TimeGAN Approach with Enhanced VaR Estimations," Jan. 2024, doi: 10.21203/RS.3.RS-3906541/V1.

[35] V. Kanaparthi, "Transformational application of Artificial Intelligence and Machine learning in Financial Technologies and Financial services: A bibliometric review," Jan. 2024, doi: 10.1016/j.jbusres.2020.10.012.

[36] V. K. Kanaparthi, "Navigating Uncertainty: Enhancing Markowitz Asset Allocation Strategies through Out-of-Sample Analysis," Dec. 2023, doi: 10.20944/PREPRINTS202312.0427.V1.

[37] V. Kanaparthi, "Robustness Evaluation of LSTM-based Deep Learning Models for Bitcoin Price Prediction in the Presence of Random Disturbances," Jan. 2024, doi: 10.21203/RS.3.RS-3906529/V1.